\documentclass[11pt,a4paper]{article}
\usepackage[hyperref]{acl2020}
\usepackage{times}
\usepackage{latexsym}

\usepackage{graphicx}
\usepackage{subfigure}
\usepackage{booktabs} 
\usepackage[none]{hyphenat}
\usepackage{bm,bbm,dsfont}
\usepackage{amsmath, amssymb}

\usepackage{pgfplots}
\pgfplotsset{compat=1.6}

\usepackage{enumitem}

\usepackage{microtype}

\aclfinalcopy 

\title{LAVA NAT: A Non-Autoregressive Translation Model with Look-Around Decoding and Vocabulary Attention}

\author{
Xiaoya Li$^{1}$,
Yuxian Meng$^{1}$, 
{Arianna Yuan}$^{1,2}$,
{Fei Wu}$^{3}$ and Jiwei Li$^1$  \\
$^1$ ShannonAI\\$^2$ Stanford University \\
$^3$ Department of Computer Science and Technology, Zhejiang University\\
 {\{xiaoya\_li, yuxian\_meng, arianna\_yuan, jiwei\_li\}}@shannonai.com, wufei@zju.edu.cn 
  }

\date{}

\begin{document}
\maketitle

\begin{abstract}
Non-autoregressive translation (NAT) models generate multiple tokens in one forward pass and is highly efficient at inference stage compared with autoregressive translation (AT) methods. However, NAT models often suffer from the multimodality problem, i.e., generating duplicated tokens or missing tokens. 
In this paper, we propose two novel methods to address this issue, the Look-Around (LA) strategy and the Vocabulary Attention (VA) mechanism. The Look-Around strategy predicts the neighbor tokens in order to predict the current token, and the Vocabulary Attention models long-term token dependencies inside the decoder by attending the whole vocabulary for each position to acquire  knowledge of which token is about to generate. 
Our proposed model uses significantly less time during inference compared with autoregressive models and most other NAT models.
Our experiments on four benchmarks (WMT14 En$\rightarrow$De, WMT14 De$\rightarrow$En, WMT16 Ro$\rightarrow$En and IWSLT14 De$\rightarrow$En) show that the proposed model achieves competitive performance compared with the state-of-the-art non-autoregressive and autoregressive  models while significantly reducing the time cost in inference phase.
\end{abstract} 

\section{Introduction}
\label{introduction}
Encoder-decoder based neural machine translation (NMT) models have achieved impressive success \citep{NIPS2014_5346,bahdanau2014neural,cho-etal-2014-learning,48560,pmlr-v70-gehring17a,vaswani2017transformer}. 
Since tokens are predicted one-by-one, this decoding strategy is called {autoregressive translation} (AT). 
Regardless of its simplicity, the autoregressive property makes the model slow to run, and thus it is often a bottleneck in parallel computing with GPUs.

Many attempts have tried to address this issue by replacing AT with non-autoregressive translation (NAT) during decoding, namely, generating multiple or even all tokens at once \citep{jiatao2018nat, xuezhe2019flowseq, lee2018iterative}. 
However, the conditional independence between target tokens in NAT makes it difficult for the model to capture the complete distribution of the target sequence.
This phenomenon is termed as the \emph{multimodality problem} \cite{jiatao2018nat}. 
The multimodality problem often causes two types of translation errors during inference \citep{yiren2019auxiliaryreg}: \emph{repeated translation}, i.e., the same token is generated repeatedly at consecutive time steps, and \emph{incomplete translation}, i.e., the semantics of several source tokens are not fully translated.

We hypothesize that this problem is related to the limitation of  NAT systems in modeling the relations between positions and tokens (the {\it position-token mismatch issue}) and the relations between target tokens (the {\it token-token independence issue}). 
For the former, current NAT approaches do not explicitly model the positions of the output words, and may ignore the reordering issue in generating output sentences \cite{bao2020pnat}.
For the latter, the independence between tokens directly results in repeated or missing semantics in the output sequence since each token has little knowledge of what tokens have been produced at other positions. 

Therefore, in this paper we propose two novel methods to address these problems: 
(1) Look-Around (LA), where for each position we first predicts its neighbor tokens on both sides and then use the two-way tokens to guide the decoding of the token at the current position; and
(2) Vocabulary Attention (VA), where each position in intermediate decoder layers attends to 
reorder prediction labels of a word based on its contextual information.  
These two methods improve the decoder in different ways. VA emphasizes the relations between tokens, whereas LA models the relations between positions and tokens. The combination of the two methods leads to more accurate translation. We also introduce a dynamic bidirectional decoding strategy to improve the translation quality with high efficiency, 
which further enhances the performance of the translation model. 
The proposed framework successfully models tokens and their orders within the decoded sequence, and thus greatly reducing the negative impact of the multimodality issue. 

We conduct experiments on four benchmark tasks, including WMT14 En$\rightarrow$De, WMT14 De$\rightarrow$En, WMT16 Ro$\rightarrow$En and IWSLT14 De$\rightarrow$En. Experimental results show that  the proposed method achieves competitive performance compared with existing state-of-the-art non-autoregressive and autoregressive neural machine translation models while significantly reducing the decoding time.
\section{Related Work}
\label{related}
Non-autoregressive (NAT) machine translation task was first introduced by \citet{jiatao2018nat} to alleviate latency during inference in autoregressive NMT systems. 
Recent work in non-autoregressive machine translation investigated the developed ways to mitigate the trade-off between decoding in parallelism and performance. 
\citet{jiatao2018nat} utilized fertility as latent variables towards solving the multimodality problem based on the Transformer network.
\citet{kaiser2018lattransformer} designed the LatentTransformer which autoencodes the target sequence into a shorter sequence of discrete latent variables and generates the final target sequence from this shorter latent sequence in parallel. 
\citet{bao2019pnat} addressed the issue of lacking positional information in NAT generation models. 
They proposed PNAT, a non-autoregressive model with modeling positions as latent variables explicitly, which narrows the gap between NAT and AT models on machine translation tasks. 
Their experiment results also indicate NAT models can achieve comparable results without using knowledge distillation and the positional information is able to greatly improve the model capability.
\citet{zhiqing2019natcrf} designed an efficient approximation for CRF for NAT models in order to model the local dependencies inside sentences. 

\section{Background}
\label{background}
Neural machine translation (NMT) is the task of generating a sentence in the target language $\bm{y}= \{y_1, y_2, ..., y_{m}\}$ given the input sentence from the source language $\bm{x}= \{x_1, x_2, ..., x_{n}\}$, where $n$ and $m$ are the length of the source and target sentence, respectively. 
\subsection{Autoregressive Neural Machine Translation}
Autoregressive decoding has been a major approach of target sequence generation in NMT. Given a source sentence $\bm{x}$ with length $n$, an NMT model decomposes the distribution of the target sentence $\bm{y}$ into a chain of conditional probabilities in a unidirectional manner
\begin{equation}
     p_\text{AT}(\bm{y}|\bm{x};\theta)=\prod_{t=1}^{m+1}p(y_t|y_{0:{t-1}},\bm{x};\theta)
\end{equation}
where $y_0$ is the special token \verb|<BOS>| (the beginning of the sequence) and $y_{m+1}$ is \verb|<EOS>| (the end of the sequence). 
{\it Beam search} \citep{wiseman-rush-2016-sequence,li-etal-2016-diversity,li2016simple} is commonly used as a heuristic search technique that explores a subset of possible translations in the decoding process. It often leads to better translation since it maintains multiple hypotheses at each decoding step.

\subsection{Non-Autoregressive Nerual Machine Translation}
Regardless of its convenience and effectiveness, the autoregressive decoding methods have two major drawbacks. One is that it cannot generate multiple tokens simultaneously, leading to an inefficient use of parallel hardwares such as GPUs. The other is that beam search has been found to output low-quality translation when applied to large search spaces \citep{koehn-knowles-2017-six}. 
Non-autoregressive translation methods could potentially address these issues. Particularly, they aim at speeding up decoding through removing the sequential dependencies within the target sentence and generating multiple target tokens in one pass, as indicated by what follows:
\begin{equation}
    p_\text{NAT}(\bm{y}|\bm{x};\phi)=\prod_{t=1}^m p(y_t|\bm{x};\phi)
\end{equation}
Since each target token $y_t$ only depends on the source sentence $\bm{x}$, we only need to apply \verb|argmax| to every time step and they can be decoded in parallel. The only challenge left is the {\it multimodality problem} \cite{jiatao2018nat}. Since target tokens are generated independently from each other during the decoding process, it usually results in duplicated or missing tokens in the decoded sequence. Improving decoding consistency in the target sequence is thus crucial to NAT models. 

\section{Method} 
\subsection{Model Architecture}
\begin{figure}
    \centering
    \includegraphics[scale=0.6]{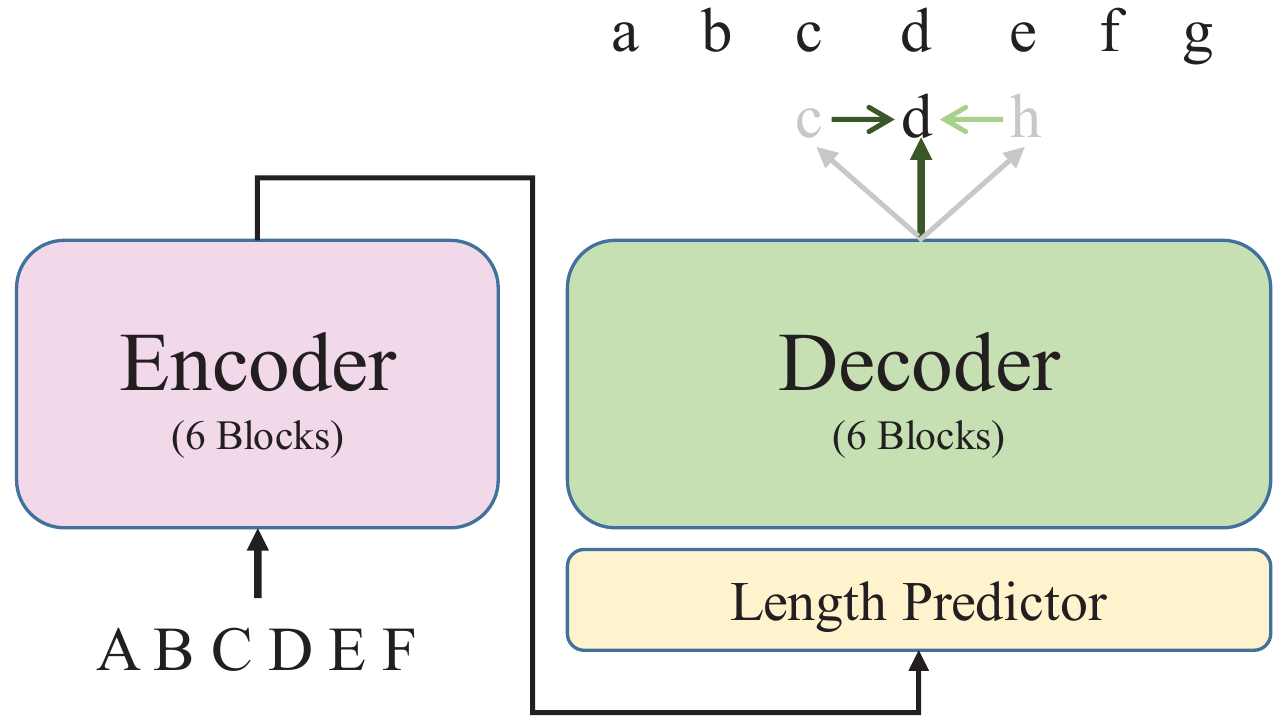}
    \caption{An overview of the proposed NAT model architecture. 
    }
    \label{fig:overview}
\end{figure}

The overview of the proposed model is shown in Figure~\ref{fig:overview}. It consists of three modules: an encoder, a target length predictor and a decoder.

\subsubsection{Encoder}
Following Transformer \cite{vaswani2017transformer}, we use a stack of $N=6$ identical transformer blocks as the encoder. Given the source sequence $\bm{x}=\{x_1,\cdots,x_n\}$, the encoder produces the contextual representations $\bm{H}=\{\bm{h}_1,\cdots,\bm{h}_n\}$, which are obtained from the last layer of the encoder.

\subsubsection{Decoder}
Following previous work \cite{jiatao2018nat,xuezhe2019flowseq,bao2019pnat}, we predict the length difference $\Delta m$ between the source and the target sequences using a classifier with output range [-20, 20]. Details are illustrated  in the supplementary material. 

The decoder also consists of 6 identical transformer blocks, but it has two key differences from the encoder. First, we introduce the vocabulary attention (VA), where each position in each decoder layer learns a vocabulary-aware representation; and (2) we propose a look-around decoding scheme (LA), in which the model first  predicts neighbor tokens on both sides of the current position and then uses the neighbor tokens to guide the decoding of the token at the current position. 

\paragraph{Input}
We adopt a simple way to feed the contextual representations (computed from the source sentence) to the decoder.  
More concretely, the decoder input $\bm{d}_i$ for the $i$-th position is simply a copy of the int($(n/m)*i$)th contextual representation, i.e., $\bm{h}_{int((n/m)*i)}$ from the encoder.

\paragraph{Positional Embeddings}
The absolute positional embeddings
in the vanilla Transformer
 may cause generation of repeated tokens or missing tokens since the positions are not modeled explicitly. We therefore use both relative and absolute positional embeddings in the NAT decoder.
For relative position information, we follow \citet{shaw2018relativepos} which learns different embeddings for different offset between the ``key'' and the ``query'' in the self-attention mechanism with a clipping distance $k$ (we set $k=4$) for relative positions.
For absolute positional embeddings, we follow \citet{radford2019gpt2} which uses a learnable positional embedding $\bm{p}_i$ for position $i$.

\paragraph{Vocabulary attention(VA)}
Although these two positional embedding strategies are integrated into the decoder, they do not fully address the issues of missing tokens and repeated tokens in decoding. This is because: (1) positional embeddings only care about ``position'' but not ``token'' themselves; (2) the relations between positions and tokens are not explicitly captured \cite{bao2019pnat}. To tackle these problems, we introduce a layer-wise vocabulary attention (VA), where each position in intermediate decoder layers attends to all tokens in the vocabulary to guess which tokens are ``ready'' to produce, and this prior information of all positions are then aggregated to model long-range dependencies during the decoding. 

More concretely, we denote the input to the decoder as $\bm{Z}^{(0)}$, which equals to $\{\bm{d}_1,\cdots,\bm{d}_m\}$ and the contextual representations in the $i$-th decoder layer as $\bm{Z}^{(i)}(1\le i\le 6)$. The intermediate  representation of position $j (1\le j\le m)$ in the $i$-th decoder layer $\bm{a}^{(i)}_j$ is thus given by:
\begin{equation}
    \bm{a}^{(i)}_j=\text{softmax}(\bm{z}^{(i)}_j\cdot\bm{W}^\mathrm{T})\cdot\bm{W}
\end{equation}
where $\bm{W}$ is the representation matrix of the token vocabulary. 
From the equation we can see that $\bm{a}^{(i)}_j$ is essentially the intermediate vocabulary representation at the $j$th position weighted by the vocabulary attention. It provides the prior information on which token is ready to be generated at each position. 
The intermediate vocabulary representations at all positions in the same layer are then aggregated to get  $\bm{A}^{(i)}=[\bm{a}^{(i)}_1,\cdots,\bm{a}^{(i)}_m]$, which captures the long-range dependencies between tokens at different positions. The representations in the next layer $\bm{Z}^{(i+1)}$ are produced by taking the concatenation of the previous layer representations and the intermediate vocabulary representations $[\bm{Z}^{(i)};\bm{A}^{(i)}]$ as input. Through this process, representations at each positions are aware of the prior information computed by the previous layer on which tokens are ready to be generated. Therefore, it helps the decoder to further narrow down to the appropriate tokens to generate. It is easy to implement and no extra parameter is introduced to the training process. 

\begin{figure}[t]
    \centering
    \includegraphics[scale=0.34]{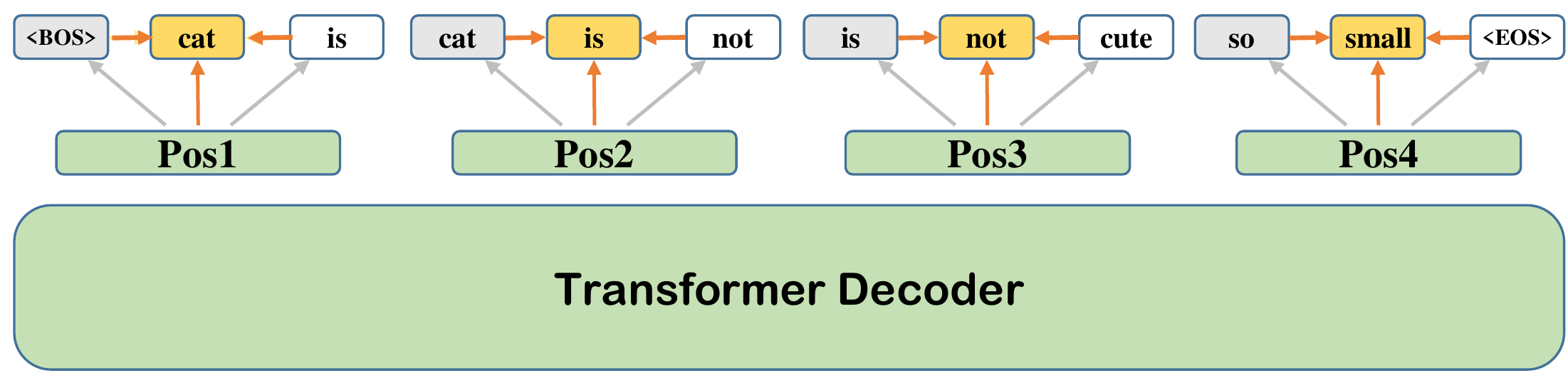}
    \caption{The overview of Look-Around Decoding. For each position, the decoder is required to first predict tokens on its left side and on its right side before generating the token at the current position. Then, the token for the current position is decoded by incorporating the tokens on both sides, with two gates controlling to what degree these tokens contribute to the current token.}
    \label{fig:la}
\end{figure}

\paragraph{Look-Around (LA) Decoding}
\label{twl}
Vocabulary attention is useful for modeling token dependencies (token-token relationship) as information passes through the decoder. However, we still need to address the position-token mismatch issue. Since the final output is decoded at the last decoder layer, we propose a {\it Look-Around}(LA) decoding scheme, where for each position, the decoder is required to first predict tokens on its left side and on its right side before generating the token at the current position. The number of predicted tokens on the left side is called left-side size (LS for abbreviation), and the size of the right side is called right-side size (RS). The LA decoding discourages predicting tokens that are the same as their neighbors, and thus is critical for modeling the position-token relation.
For ease of exposition, we use LS=RS=1 for illustration. The influence of LS and RS will be discussed in experiments. Figure~\ref{fig:la} illustrates the Look-Around mechanism.

Formally, for position $i$, the optimization objective is changed from $p(y_i|\bm{x};\phi)$ to $p(y_{i,l}|\bm{x},\bm{p}_{i-1};\phi)$, $p(y_i|\bm{x},\bm{p}_i;\phi)$ and $p(y_{i,r}|\bm{x},\bm{p}_{i+1};\phi)$, given by:
\begin{gather}
    p(y_{i,l}|\bm{x},\bm{p}_{i-1};\phi)=\text{softmax}(W_l(\bm{z}_i+\bm{p}_{i-1})+b_l)\\
    p(y_{i,r}|\bm{x},\bm{p}_{i+1};\phi)=\text{softmax}(W_r(\bm{z}_i+\bm{p}_{i+1})+b_r)\\
    p(y_{i}|\bm{x},\bm{p}_{i};\phi)=\text{softmax}(W\bar{\bm{z}}_i+b)\\
    \bar{\bm{z}}_i=
    (\bm{z}_i+\bm{p}_i)\oplus(\bm{c}_{i,l}\odot\bm{w}_{i,l})
    \oplus(\bm{c}_{i,r}\odot\bm{w}_{i,r})
\end{gather}
where $\bm{p}$ is the learnable absolute positional embedding, $\bm{z}_i$ is the $i$-th vocabulary attention representation from the decoder's last layer ({\it i.e.} $\bm{Z}=\bm{A}^{(6)}$), $\oplus$ denotes the concatenation operation and $\odot$ denotes  element-wise multiplication. For each position $i$, the decoder first predicts its left-sided tokens $x_{i,l}$ and its right-sided tokens $x_{i,r}$ through Eq.~(4) and (5), respectively. The word embeddings with respect to $x_{i,l}$ and $x_{i,r}$ are denoted as $\bm{w}_{i,l}$ and $\bm{w}_{i,r}$. 
Then, the two-way predictions are aggregated to predict the token at position $i$ through two gates $\bm{c}_{i,l}$ and $\bm{c}_{i,r}$, which control what information of the neighbor tokens should be considered when predicting $y_i$ (Eq.~(6) and (7)). The gate states are calculated by: 
\begin{equation}
\begin{aligned}
    \bm{c}_{i,l}&=\sigma(W'_{l}(\bm{w}_{i,l}+\bm{p}_{i-1})+b'_l)\\
    \bm{c}_{i,r}&=\sigma(W'_{r}(\bm{w}_{i,r}+\bm{p}_{i+1})+b'_r)
\end{aligned}
\end{equation}
where $\sigma$ is the sigmoid function. Note that this process is not sequential and thus can be implemented in a parallel manner. Particularly, it requires ``one-pass'' to get the representation at each position $\bm{z}_i$, and running the neighbor-token readout classifier for each location in parallel and running the current-token readout classifier for each location in parallel.

\subsection{Training}

We use $(\bm{x},\bm{y}^*)$ to denote a training instance where $\bm{x}$ is the input source sequence and $\bm{y}^*=\{y^*_1,\cdots,y^*_{m'}\}$ is the target sequence with length $m'$. The decoded sequence is ${\bm{y}}=\text{Decoder}(\text{Encoder}(\bm{x}))=\{y_1,\cdots,y_{m}\}$. 

Differentiable scheduled sampling  \citet{goyal2017differentiable} is  used to address the exposure bias issue\footnote{which refers to the train-test discrepancy that arises when an autoregressive generative model uses only ground-truth contexts during training but generated ones at test time.}:
 a {peaked softmax function}  is used to define a differentiable soft-argmax procedure: 
\begin{equation}
\overline{\bm{w}}_{i-1} = \sum_{y} {\bm{w}}(y) \cdot \frac{\text{exp}(\alpha s_{i-1}(y))}{\sum_{y_{'}} \text{exp}(\alpha s_{i-1}(y_{'}) ) }
\end{equation}
where $s_{i-1}(y)$ is the logit calculated at position $i-1$.
When $\alpha \rightarrow \infty$, the equation above approaches the argmax operation. With a large yet finite $\alpha$, we get a linear combination of all the tokens that are dominated by the one with the maximum score.
We replace $\bm{w}_{i, l}$ and $\bm{w}_{i, r}$ in Eq.~(7) and (8) with $\overline{\bm{w}}_{i, l}$ and $\overline{\bm{w}}_{i, r}$ during training.

\paragraph{Loss Function} 
During training, we propose to combines the bag-of-words loss $L_\text{bow}$ \cite{shuming2018bagofwords} with the standard CE loss. 
The bag-of-words loss 
used the bag-of-words as the training target to encourage the model to generate potentially correct sentences that did not appear in the training set. 
The bag-of-words loss is computed by: 
\begin{gather} 
    p_\text{bow}=\text{sigmoid}\left(\sum_{i=1}^m(W\bar{\bm{z}}_i+b)\right)\\
    \mathcal{L}_\text{bow} = - \sum_{i=1}^{|V|}\mathds{1}\{v_i\in\bm{y}^*\}\log p_\text{bow}(v_i) 
\end{gather} 
where $p_\text{bow}$ is a vector of length $|V|$ and $V$ is the vocabulary of the target language. Each value in $p_\text{bow}$ represents the probability of how possible the word appears in the generated sentence regardless of its position.
This gives the final loss as follows:
\begin{gather}
    \mathcal{L}_\text{ce}=-\sum_{i=1}^{m'}\log p(y^*_i)\\ 
    \mathcal{L} = \mathcal{L}_\text{ce} + \lambda_{t} \mathcal{L}_\text{bow}
\end{gather} 
$\lambda_{t}$ is the coefficient to balance the two loss functions at the $t$-th epoch. 
Hyperparameters $\lambda, k, \alpha$ are tuned using the validation set.

\begin{figure*}
    \centering
    \small
    \includegraphics[scale=0.45]{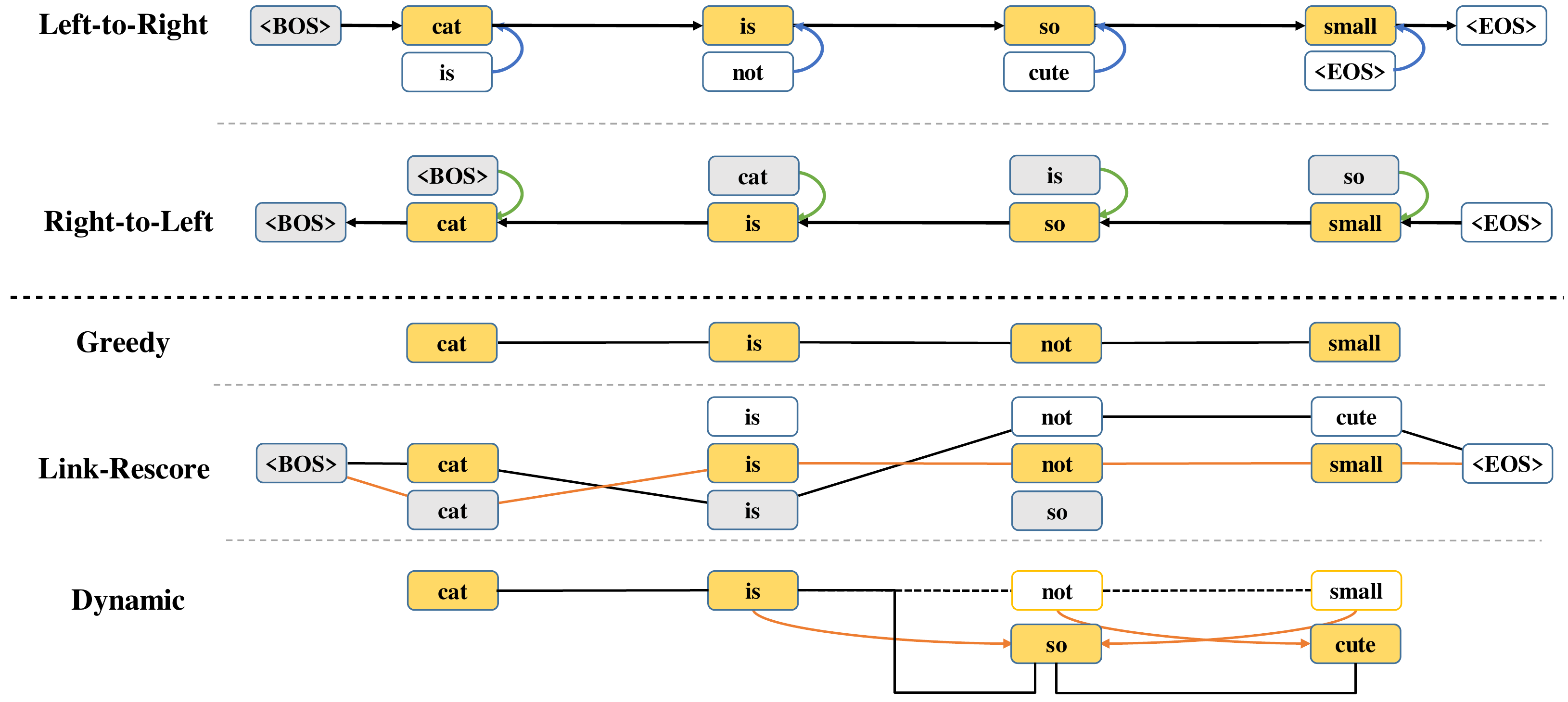}
    \caption{Illustration of different decoding strategies. Note that the sequential decoding does not use the Look-Around mechanism.}
    \label{fig:decode}
\end{figure*}

\subsection{Inference}
In this subsection, we discuss three decoding algorithms for the proposed model: \emph{partially sequential decoding},  
\emph{static decoding}, and \emph{dynamic decoding}. 
Fig.~\ref{fig:decode} gives an illustration for these decoding methods.
Static and dynamic decoding are based on the look-around (left\&right) decoding mechanism, whereas partially sequential decoding is based on one-side (left/right) mechanism.  

\subsubsection{Partially Sequential Decoding} 
Partially sequential decoding is based on part of look-around mechanism and tokens are generated in the left-to-right or right-to-left manner. 
We assume the target sequence length is $m$ and take left-to-right decoding as an example. 
For position $i$, the model generates its  right-side token $y_{i,r}$ and the current token $y_{i}$ is generated depending on $y_{i,r}$ and the previous generated token $y_{i-1}$. 
A sequence of tokens will be obtained after repeating this process $m$ times. 
We also apply beam search for finding approximately-optimal output translations.  
%


\subsubsection{Static Decoding}


\paragraph{Greedy Decoding} 
For position $i$, greedy decoding removes $y_{i,l}$ and $y_{i, r}$ from $(y_{i,l}, y_{i}, y_{i, r})$ triples and selects the token with the the highest probability in Eq.~(6). 
The process generates one sequence of tokens as the output. 

\paragraph{Noisy Parallel Decoding}
We follow the noisy parallel decoding strategy proposed in \citet{jiatao2018nat}, which generates a number of decoding candidates with various lengths in parallel. 
After collecting multiple translation results, 
we use our pretrained teacher model to evaluate the results and select the one that achieves
the highest probability.
The rescoring operation does not hurt the non-autoregressive property of the LAVA model. 

\paragraph{Link and Rescore}
Link-rescore collects tokens from triples and rearrange them in order. 
Link-rescore treats $y_{i-1, r}$, $y_{i}$, $y_{i+1, l}$ as candidates for filling in the blank at position $i$ of the output. 
A number of decoding candidates are generated after randomly selecting one token from three candidates to fill in the blank for each  timestep.  
Afther that, we use pretrained AT models to re-score all these candidate sequences and choose the one with the highest probability as the final output. 


\subsubsection{Dynamic Bidirectional Decoding}
Dynamic bidirectional decoding is based on the look-around (LA) scheme. 
We first obtain an initial sequence in the same way as we do in greedy decoding. Then 
we continue refining this sequence by masking out and re-predicting a subset of tokens whose probability is under a threshold in the current translation. 
The masked token $y_j$ is then re-predicted by depending on the initial sequence predictions $y_{j-1}$ and $y_{j+1}$. 
The final output is obtained after $R$ times refinement and each token can regenerate at most once. 


As shown in Fig.~\ref{fig:decode}, the original translation is '\emph{cat is not small}', where the probabilities of tokens ``not'' and ``small'' are under the threshold {$p$}, so the tokens on position 3 and position 4 are re-predicted by ``is'', ``small'' and by ``not'', respectively. The re-predicted tokens are now ``so'' and ``cute'', the current translation is thus '\emph{cat is so cute}'. We set the maximum number of refinement turns $R$ to {4} for all our experiments.

\begin{table*}[t]
    \small
    \centering
    \scalebox{0.85}{
      \setlength{\tabcolsep}{4pt}
      \begin{tabular}{l@{\hspace{0.3cm}}c@{\hspace{0.25cm}}c@{\hspace{0.25cm}}c@{\hspace{0.25cm}}c@{\hspace{0.25cm}}c@{\hspace{0.25cm}}c@{\hspace{0.25cm}}c@{\hspace{0.25cm}}c@{\hspace{0.25cm}}}
      \toprule
       & \multicolumn{1}{c}{\bf  WMT14 En$\rightarrow$De}& \multicolumn{1}{c}{\bf WMT14  De$\rightarrow$En}& \multicolumn{1}{c}{\bf WMT16 Ro$\rightarrow$En}&
       \multicolumn{1}{c}{\bf IWSLT14 De$\rightarrow$En}& 
        \multicolumn{1}{c}{\bf } & 
         \multicolumn{1}{c}{\bf }
       \\\hline
      {\bf \emph{Autoregressive Models}} &   {\bf }  & {\bf } & {\bf}  & {\bf } & {\bf latency} & {\bf speedup } \\
      \midrule
      LSTM Seq2Seq \citep{bahdanau2016actor} & 24.60 & - & - & 28.53 &-& -  \\
      ConvS2S \citep{edunov2017classical} &26.43 & - & - & 32.84 &-&-\\
      Transformer \citep{vaswani2017transformer}  & 27.30& 31.29& -& 33.26 & 784 ms & 1.00x \\\midrule
      {\bf \emph{Non-autoregressive Models}} &   {\bf }  & {\bf } & {\bf}  & {\bf } & {\bf latency} & {\bf speedup } \\\midrule
      NAT \citep{jiatao2018nat} & 17.69  & 20.62 & 29.79 & - & 39 ms & 15.6x   \\
       NAT (rescore 10 candiates) & 18.66  & 22.41 & - & - &  79 ms & 7.68 x   \\
      NAT (rescore 100 candiates) & 19.17  & 23.20 & - & - &  257 ms & 2.36 x   \\
      iNAT \citep{lee2018iterative} & 21.54  & 25.43 & 29.32 & - &  - & 5.78 x   \\
      Hint-NAT \citep{zhuohan2019hint} & 21.11  & 25.24 & - & 25.55 &  26 ms & 30.2 x   \\
      Hint-NAT (beam=4,rescore 9 candidates) & 25.20  & 29.52 & - & 28.80 &  44 ms & 17.8 x   \\
      CRF-NAT \citep{zhiqing2019natcrf} & 23.32  & 25.75 & - & 26.39 &  35 ms & 11.1 x   \\
      CRF-NAT (rescore 9 candidates) & 26.04  & 28.88 & - & 29.21 &  60 ms & 6.45 x   \\
      DCRF-NAT \citep{zhiqing2019natcrf} & 23.44  & 27.22 & - & 27.44 &  37 ms & 10.4 x   \\
      DCRF-NAT (rescore 9 candidates) & 26.07  & 29.68 & - & 29.99 &  63 ms & 6.14 x   \\
      DCRF-NAT (rescore 19 candidates) & \underline{26.80}  & \underline{30.04} & - & 30.36 &  88 ms & 4.39 x   \\
      FlowSeq-base (raw data) \citep{xuezhe2019flowseq}  & 18.55  & 23.36 & 29.26 & 24.75 &  - & -   \\
      FlowSeq-base (kd) \citep{xuezhe2019flowseq}  & 21.45  & 26.16 & 29.34 & 27.55 &  - & -  \\
      FlowSeq-large (raw data) \citep{xuezhe2019flowseq}  & 20.85  & 25.40 & \underline{29.86} & - &  - & -   \\
      FlowSeq-large (kd) \citep{xuezhe2019flowseq}  & 23.72  & 28.39 & 29.73 & - &  - & -  \\
      PNAT \citep{bao2019pnat} & 19.73  & 24.04 & - & - &  - & -  \\
      PNAT (kd) \citep{bao2019pnat} & 23.05  & 27.18 & - & - &  - & -  \\
      PNAT (best) \citep{bao2019pnat} & 24.48  & 29.16 & - & \underline{32.60} &  - & 3.7 x  \\\midrule
      {\bf LAVA NAT (w/ kd)} & 25.29  & 30.21 & 32.11 & 30.73 &  28.32 ms & 29.34 x   \\ %
      {\bf LAVA NAT (w/ kd, rescore 10 candidates)} & 26.89  & 30.72 & 32.38 & 31.64 &  57 ms & 6.56 x   \\
      {\bf LAVA NAT (w/ kd, rescore 20 candidates)} & 27.32  & {\bf 31.46} & 32.64 & 31.51 &  93 ms & 3.91 x   \\
      {\bf LAVA NAT (w/ kd, dynamic decode, rescore)} & {\bf 27.94}  & 31.33 & {\bf 32.85} & {\bf 31.69} &  34.29 ms & 20.18 x   \\\midrule
      {\bf LAVA NAT (w/o kd)} & 25.72  & 30.04 & 32.26 & 30.36 &  28.32 ms & 29.34 x   \\ %
      {\bf LAVA NAT (w/o kd, rescore 10 candidates)} & 27.89  & 31.24 & 32.63 & 31.57 &  57 ms & 6.56 x   \\
      {\bf LAVA NAT (w/o kd, rescore 20 candidates)} & 27.93  & {\bf 31.59} & 32.40 & 33.42 &  93 ms & 3.91 x   \\
      {\bf LAVA NAT (w/o kd, dynamic decode, rescore)} & {\bf 28.42}  & 31.46 & {\bf 32.93} & {\bf 33.59} &  33.84 ms & 20.18 x    \\
      \bottomrule\hline
      
      \end{tabular}
      }
    \caption{The performances of non-autoregressive methods on WMT14 En$\leftrightarrow$De, WMT16 Ro$\rightarrow$En and IWSLT14 De$\rightarrow$En datasets in terms of {BLEU score} .  We use "kd" to denote "knowledge distillation". "best" represents the highest BLEU score reported in the original paper.}
    \label{overall-performance}
    \vskip -0.15in
    \end{table*}

\section{Experiments}
\subsection{Experiment Settings}
We evaluate the proposed method on four machine translation benchmark tasks (three  datasets):
WMT2014 De$\rightarrow$En (4.5M sentence pairs), WMT2014 En$\rightarrow$De, WMT2016 Ro$\rightarrow$En (610K sentence pairs) and IWSLT2014 De$\rightarrow$En (150K sentence pairs). We use the Transformer \citep{vaswani2017transformer} as a backbone.  
{\bf Knowledge Distillation} is applied for all models. 
The details of the model setup and Knowledge Distillation are presented in the supplementary material.

We evaluate the models using tokenized case-sensitive BLEU score \citep{papineni2002bleu} for WMT14 datasets and tokenized case-insensitive BLEU score for IWSLT14 datasets. Latency is computed as the average decoding time (ms) per sentence on the full test set without mini-batching. 
The training and decoding speed are measured on a single NVIDIA Geforce GTX TITAN Xp GPU. 

\paragraph{Baselines} 
We choose the following models as baselines:
\begin{itemize}[topsep=0pt, partopsep=0pt] 
\setlength{\itemsep}{0pt}%
\setlength{\parskip}{0pt}%
\item {\bf NAT:}  The  NAT model introduced by \citet{jiatao2018nat}. 
\item {\bf iNAT:} \citet{lee2018iterative} extended the vanilla NAT model by iteratively reading and refining the translation. The number of iterations is set to 10 for decoding. 
\item {\bf FlowSeq:} \citet{xuezhe2019flowseq} adopted normalizing flows \citep{kingma2018glow} as latent variables for genernation. 
\item {\bf Hint-NAT:} \citet{zhuohan2019hint} utilized the intermediate hidden states from an autoregressive  teacher  to improve the NAT model. 
\item {\bf CRF-NAT/DCRF-NAT:} \citet{zhiqing2019natcrf} designed an approximation of CRF for NAT models (CRF-NAT) and further used a dynamic transition technique to model positional contexts in the CRF (DCRF-NAT). 
\item {\bf PNAT:} \citet{bao2020pnat} proposed a non-autoregressive transformer architecture by position learning (PNAT), which explicitly models the positions of output words as latent variables during text generation.
\end{itemize} 

\subsection{Results}
Experimental results are shown in Table \ref{overall-performance}. We first compare the proposed method against the autoregressive counterparts in terms of generation quality, which is measured by BLEU \citep{papineni2002bleu}. 
For all our tasks, we obtain results comparable with Transformer, the state-of-the-art autoregressive model. Our best model achieves 28.42 (+1.1 gain over Transformer), 31.59 (+0.3 gain) and 33.59 (+0.3 gain) BLEU score on WMT14 En$\rightarrow$De, WMT14 De$\rightarrow$En and IWSLT14 De$\rightarrow$En, respectively. More importantly, our LAVA-NAT model decodes much faster than Transformer, which is a big improvement regarding the speed-accuracy trade-off in AT and NAT models.

Comparing our models with other NAT models, we observe that the best LAVA-NAT model achieves significantly huge performance boost over NAT, iNAT, Hint-NAT, DCRF-NAT, FlowSeq and PNAT by +9.25, +6.88, +3.22, +1.62, +4.7 and +3.94 in BLEU on WMT14 En$\rightarrow$De, respectively. It indicates that the Look-Around decoding strategy greatly helps reducing the impact of the multimodality problem and thus narrows the performance gap between AT and NAT models. In addition, we see a +1.55, +3.07 and +0.99 gain of BLEU score over the best baselines on WMT14 De$\rightarrow$En, WMT16 Ro$\rightarrow$En and IWSLT14 De$\rightarrow$En, respectively. 

From the last two groups in Table~\ref{overall-performance}, we find that the rescoring technique helps a lot for improving the performance, and dynamic decoding significantly reduces the time spent on rescoring while further enhancing the decoding process. On WMT14 En$\rightarrow$De, rescoring 10 candidates leads to a gain of +1.6 BLEU, and rescoring 20 candidates gives about a +2 BLEU score increase. The same phenomenon can be observed on other datasets and for models without knowledge distillation. Nonetheless, rescoring has the weakness of greater time cost, for which dynamic decoding is beneficial to mitigate this issue by speeding up about 3x$\sim$5x comparing to decoding with AT teacher rescoring mechanism.

\subsection{Decoding Speed}
We can see from Table~\ref{overall-performance} that LAVA NAT gets a nearly 30 times decoding speedup than Transformer, while achieving comparable performances. Compared to other NAT models, we can observe that the LAVA NAT (w/kd) model is almost the fastest (only a little bit behind of Hint-NAT) in terms of latency, and is surprisingly faster than CRF-NAT and PNAT, with better performances. In addition, it's worth noting that dynamic decoding is vital in speeding up decoding (3$\sim$5 times faster than LAVA NAT with rescoring) while achieving promising results. 

\begin{table}[t]
    \small
    \center
    \begin{tabular}{lcc}\toprule
    \multicolumn{3}{c}{{\bf IWSLT14 De$\rightarrow$En}}\\\midrule
    {\bf Model} &  {\bf origin}  & {\bf origin+BOW} \\\midrule
    NAT & 23.67 & 24.17(+0.50) \\ 
    FlowSeq-base  & 27.55 & 28.50(+0.95)  \\
    DCRF-NAT  & 29.99 & 30.14(+0.15) \\
    Hint-NAT & 25.55 & 25.85(+0.30) \\ \bottomrule
    \end{tabular}
    \caption{The effect of using Bag-of-Words as targets.}
    \label{bag-of-words}
    \end{table}
\begin{table}[t]
    \small
    \center
    \scalebox{0.9}{
    \begin{tabular}{ccc}\toprule
    {\bf Method} &  {\bf WMT14 En$\rightarrow$De} & {\bf IWSLT14 De$\rightarrow$En}    \\\midrule
     w/ VA    & 25.72 & 30.36  \\ 
    w/o VA & 24.62 & 28.80 \\\bottomrule
    \end{tabular}
    }
    \caption{The effect of vocabulary attention(VA).}
    \label{effect-softmax}
    \end{table}

\begin{table}[t]
    \small
    \center
    \begin{tabular}{lll}\toprule
    {\bf Model} &  {\bf WMT14 En$\rightarrow$De} & {\bf IWSLT14 De$\rightarrow$En}    \\\midrule
    RS=0 LS=0 & 19.35 & 23.57  \\ 
    RS=0 LS=1 & 23.94(+4.59) & 27.89(+4.32) \\ 
    RS=1 LS=0 & 22.82(+3.47) & 26.90(+3.33) \\
    RS=1 LS=1 & 28.43(+9.08) & 28.78(+5.21) \\ \bottomrule
    \end{tabular}
    \caption{The effect of changing look around (LA) size.}
    \label{effect-of-twl}
    \end{table}

\section{Analysis and Ablations}
\label{experiment-discussion}

\paragraph{Effect of Bag-of-Words Loss}
We conduct experiments on IWSLT2014 De$\rightarrow$En to verify its effectiveness. 
Experiment results are listed in Table \ref{bag-of-words}. 
Incorporating various NAT models with ``+BOW'' achieve consistent improvement in terms of BLEU score and this is in line with our expectation that sentence-level training objective can help to further boost the performance.

\paragraph{Effect of Vocabulary Attention(VA)}
We conduct experiments on our LAVA NAT model on WMT14 En$\rightarrow$De and IWSLT14 De$\rightarrow$En to show the effectiveness of the LITA mechanism.
Experimental results are shown in Table~\ref{effect-softmax}. It demonstrates that incorporating VA can help improve the translating performance (+1.1 on WMT14 En$\rightarrow$De and +1.56 on IWSLT14 De$\rightarrow$En).
This is in line with our expectation that aggregating layer-wise token information in intermediate layers can help improve the decoder's ability of capturing token-token dependencies .

\paragraph{Effect of Look Around (LA) Size}
We investigate  the effect of the LA size, {\it i.e.} RS(right size) and LS(left size) (Section~\ref{twl}). We respectively consider $\text{RS}=0,1$ and $\text{LS}=0,1$ for comparison.
We evaluate the LAVA NAT model on the IWSLT2014 En$\rightarrow$De development set. When training, we tune the hyperparameters  for every setting to obtain the optimal performance. The results are listed in Table \ref{effect-of-twl}. Surprisingly huge performance gains are obeserved by utilizing look-around. On WMT14 En$\rightarrow$De, simply setting RS=1 or LS=0 obtains a gain of +3.5$\sim$+4.6 BLEU score, and using both sides leads to a gain of +9 BLEU socre. For IWSLT14 De$\rightarrow$En, the gain is +5.21 BLEU score.

\paragraph{Effect of the Proposed Method for Other NAT Models}
In order to further test the effectiveness of the proposed look around(LA) mechanism, we incorporate LA into other NAT models on the IWSLT2014 De$\rightarrow$En dataset. 
Experimental results are shown in Table \ref{twl-iwslt14}. 
The LA mechanism succeeds in modeling local word orders inside the sentence, resulting in better translating performances.

\begin{table}[t]
    \small
    \center
    \begin{tabular}{lll}\toprule
    \multicolumn{3}{c}{{\bf IWSLT 2014 De-En }}\\\midrule
    {\bf Model} &  {\bf origin}  & {\bf origin+LA}   \\\midrule
    NAT & 23.67 & 24.58(+0.91) \\ 
    FlowSeq-base & 27.55 & 28.43(+0.88)  \\
    DCRF-NAT   & 29.99 & 30.54(+0.55) \\
    Hint-NAT & 25.55 & 26.71(+1.16) \\ 
    \bottomrule
    \end{tabular}
    \caption{BLEU score on IWSLT2014 En$\rightarrow$De translation task. We apply look-around(LA) for each NAT model to show its effectiveness.
    }
    \label{twl-iwslt14}
\end{table}

\paragraph{Effect of Differentiable Scheduled Sampling}
\begin{table}[t]
    \small
    \center
    \begin{tabular}{ccc}\toprule
    {\bf Method} &  {\bf WMT14 En$\rightarrow$De} & {\bf IWSLT14 De$\rightarrow$En}    \\\midrule
    TF  & 28.01 & 28.66  \\ 
    SS & 28.25 & 28.69 \\ 
    DSS & 28.43 & 28.78 \\ \bottomrule
    \end{tabular}
    \caption{The effect of different sampling mechanisms. ``TF'', ``SS'' and ``DSS'' are short for ``Teacher Forcing'', ``Scheduled Sampling'' and ``Differentiable Scheduled Sampling'', respectively.}
    \label{effect-of-sample}
    \end{table}

We 
conduct ablation experiments on  WMT14 En$\rightarrow$DE and IWSLT14 De$\rightarrow$EN with the LAVA NAT model. 
Experimental results are shown in Table~\ref{effect-of-sample}. Teacher forcing gets the worst performance while differentiable scheduled sampling the best and scheduled sampling is slightly better than teacher forcing. We believe this is the differentiable nature of DSS that makes the model easy to learn and thus alleviate the ``exposure bias'' issue.


\paragraph{Effect of Decoding Strategies}
\begin{table}[t]
    \center
    \small
    \scalebox{0.8}{
    \begin{tabular}{lccc}\toprule
    {\bf Model} &  {\bf W'14 En$\rightarrow$De} & {\bf I'14 De$\rightarrow$En}   & {\bf speedup} \\\midrule
    {\bf Static Decoding} \\\midrule
    Greedy &  25.72 & 30.36 & 29.34 x  \\ 
    NPD(rescore 10) & 27.89 &  31.57 & 6.56 x   \\
    Link\&Rescore & 27.54 & 31.28 &    4.74 x\\ \midrule
    {\bf Sequential Decoding} \\ \midrule
    L2R(beam=4) & 25.92 & 30.83 &  14.35 x \\ 
    L2R(beam=10) & 26.18 & 31.44 & 3.86 x \\
    R2L(beam=4) & 25.47 & 30.91 & 14.35 x \\ 
    R2L(beam=10) & 26.20 & 31.37 & 3.86 x \\\midrule
    {\bf Dynamic Decoding} \\\midrule
    Dynamic Decoding  &  28.42 & 33.59 & 20.18 x \\\bottomrule
    \end{tabular}
    }
    \caption{The effect of different decoding strategies. }
    \label{effect-decode}
\end{table}
We conduct comparative experiments on WMT14 En$\rightarrow$De and IWSLT14 De$\rightarrow$En datasets to analyze different decoding strategies. 
As shown in Table \ref{effect-decode}, {Dynamic Bidirectional Decoding} achieves the highest BLEU score across two datasets and speeds up inference by 20.18 x faster than Transformer at the same time. 
We also observe that L2R and R2L outperform greedy decoding, which proves that right and left tokens help mitigate the repeated generation problem.

\begin{figure}[h]
\begin{tikzpicture}
\begin{axis}[
width=1\columnwidth,
height=0.65\columnwidth,
xlabel={Sentence Length},
ylabel={BLEU},
grid=major,
legend entries={Transformer,NAT,LAVA},
legend style={at={(axis cs:30,23)},anchor=north west,font=\small},
]
\addplot 
table
{
 X Y
 5 28.21
 15 28.06
 25 27.89
 35 27.65
 45 24.47
};
\addplot 
table
{
 X Y
 5 20.73
 15 20.25
 25 19.36
 35 16.80
 45 16.35
};
\addplot 
table
{
 X Y
 5 27.83
 15 27.60
 25 26.51
 35 25.94
 45 25.79
};
\end{axis}
\end{tikzpicture}
\caption{Performances with respect to different sentence lengths.}
\label{fig:length}
\end{figure}
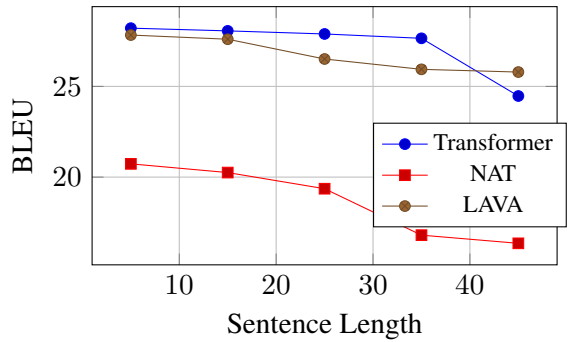

\paragraph{Effect of Sentence Length}
To evaluate different models on different sentence lengths, we conduct experiments on the  WMT14 En$\rightarrow$De development set and divide the sentence
pairs into different length buckets according to the length of the reference sentences. Figure~\ref{fig:length} shows our results.
We can see that the performance of NAT-Base drops
quickly as sentence length increases, while Transformer and the proposed LAVA-NAT models have stable performances
over different sentence lengths. This result proves the power of the proposed model in modeling long-term token dependencies.

\paragraph{Examples}
We put translation examples in  the supplementary material. 
As can be seen, 
for the NAT model, the same token is generated repeatedly at consecutive time steps. Comparing to the output of NAT, the output of  LAVA NAT does not contain repetition and more faithful to the source sentence.
The translating sentences generated by the LAVA NAT always keeps the high consistency with the reference.


\section{Conclusion}
\label{conclusion}
In this paper, we propose 
look-around decoding and vocabulary attention strategies 
for non-autoregressive NMT models. 
The proposed NAT model is one-pass, greatly reducing the time cost for inference compared to AT models and most other NAT models.
Experiments show that our method gains significantly huge performance improvements against existing state-of-the-art NAT models and competitive results to AT models.

\bibliography{acl2020}
\bibliographystyle{acl_natbib}
\newpage
\appendix

\section{Target Length Predictor}
Following previous work \cite{jiatao2018nat,xuezhe2019flowseq,bao2019pnat},
the difference $\Delta m$ between the source and the target sequences is predicted using a classifier with output range [-20, 20]. This is accomplished by first applying max-pooling to the source embeddings to get a single vector and then feeding this vector to a linear layer followed by a softmax operation:
\begin{equation}
    p(\Delta m|\bm{x})=\text{softmax}(W_{p}(\text{maxpool}(\bm{H}))+b_p)
\end{equation}
This classifier is trained jointly with the other parts of the model. In the following sections, we assume the length of the source sequence is $n$ and the predicted target length is $m$, which equals to $n+\Delta m$.
\section{Model Setup}
We follow the hyperparameters from the original Transformer-Base model \citep{vaswani2017transformer}: 6 layers for both the encoder and the decoder; 8 attention heads; 512 model dimensions and 2048 hidden dimensions. The layer normalization parameters are initialized as $\beta=0$, $\alpha=1$. 
For all experiments, we tune the dropout rate in the range of $\{0.1, 0.2, 0.3\}$ on the validation set. We also use 0.01 $L_2$ weight decay and label smoothing with $\gamma=0.1$. 
Each mini-batch contains 128K tokens and we choose Adam \citep{kingma2014adam} as the optimizer with $\beta=\{0.9, 0.999\}$ and $\gamma=0.1$. 
The learning rate warm up is $5\cdot 10^{-4}$ in the first 10K steps, and then decays under the inverse square-root schedule. 
For each model, we average the 5 best checkpoints to obtain the final results. 
\section{Knowledge Distillation}
Following previous work \citep{jiatao2018nat, chunting2019distillation}, we also use a {sequence-level knowledge distillation} from autoregressive machine translation models. We use Transformer-Base as the teacher model. 
We also initialize the encoder in the NAT models with the weights from its teacher model to enhance the translation performance \citet{jiatao2018nat, jiatao2019levenshtein}.

\begin{table*}[t]
    \small
    \centering
    \setlength{\tabcolsep}{4pt}
    \begin{tabular}{l@{\hspace{0.3cm}}l@{\hspace{0.25cm}}}
    \toprule
    \emph{Source:} & jeden morgen fliegen sie 240 kilometer zur farm . \\
    \emph{Target:} & every morning , they fly 240 miles into the farm .  \\
    \emph{AT:}  &  every morning , they fly 240 miles to the farm .  \\ 
    \emph{NAT: } &  every morning , you fly 240 miles to every morning . \\
    \emph{LAVA NAT:}  &  every morning , they fly 240 miles into the farm .  \\\midrule
    \emph{Source:} & und manches davon hat funktioniert und manches nicht . \\
    \emph{Target:} & and some of it worked , and some of it didn \&apos;t . \\
    \emph{AT: } & and some of it worked and some of it didn \&apos;t work .\\
    \emph{NAT:} & and some of it worked . \\ 
    \emph{LAVA NAT:}  &and some of it worked , and some of it didn . \\\midrule
     \emph{Source: } &  aber bei youtube werden mehr als 48 stunden video pro minute hochgeladen .\\
    \emph{Target: }  & but there are over 48 hours of video uploaded to youtube every minute . \\
    \emph{AT: } & but on youtube , more than 48 hours of video are uploaded per minute. \\
    \emph{NAT: } & but youtube , more than 48 minute of video are uploaded per hour. \\
    \emph{LAVA NAT: } &  but there are more over 48 hours of video uploaded to youtube every minute . \\\midrule 
    \emph{Source: }  & bei der coalergy sehen wir klimaveranderung als eine ernste gefahr fur unser geschaft. \\
    \emph{Target: }  & at coalergy we view climate change as a very serious threat to our business . \\
    \emph{AT: }  &  in coalergy , we see climate change as a serious threat to our business . \\
    \emph{NAT: }  &  in the coalergy , we \&apos;ll see climate climate change change as a most serious danger
    for our business . \\ 
    \emph{LAVA NAT: }  &  at coalergy , we can seeing climate change as a serious threat to our business .\\\midrule 
    \emph{Source: } & ich weiß , dass wir es können , und soweit es mich betrifft ist das etwas , was die welt jetzt braucht . \\
    \emph{Target: }  &  i know that we can , and as far as i 'm concerned , that 's something the world needs right now . \\ 
    \emph{AT: } & i know that we can , and as far as i 'm concerned , that 's something that the world needs now  \\
    \emph{NAT: } & i know that we can it , , as as as as it it it is , it 's something that the world needs now . \\
    \emph{LAVA NAT: } & i know we can do it , and as far as i 'm concerned that 's something that the world needs now . \\\midrule 
   
    \emph{Source: } &   dies ist die großartigste zeit , die es je auf diesem  
     planeten gab , egal , welchen maßstab sieanlegen \\ 
     & :gesundheit , reichtum , mobilitat ,  gelegenheiten , sinkende krankheitsraten . \\
    \emph{Target: } & this is the greatest time there \&apos;s ever been on this  planet by any measure \\
    & that you wish to choose : health , wealth , mobility , opportunity , declining rates of disease . \\
    \emph{AT: } & this is the greatest time you \&apos;ve ever had on this planet , no matter what scale you \\
    &  \&apos;re putting : health , wealth , mobility , opportunities , declining disease rates . \\
    \emph{NAT: } & this is the most greatest time that ever existed on this planet no matter what scale they \\
    & \&apos;re imsi : : , , mobility mobility , , scaniichospital rates . \\
    \emph{LAVA NAT: } & this is the greatest time that we \&apos;ve ever been on this planet no matter what scale they \\
    & \&apos;re ianition : health , wealth , mobility , opportunities , declining disease rates .\\\midrule 
    \emph{Source:}  & jetzt appelliert airbus vor der dubai airshow , wo die 777x mit über 100 bestellungen voraussichtlich das \\
    & rennen machen wird , direkt an die öffentlichkeit . \\
    \emph{Target:} & now , airbus is appealing directly to the public ahead of the dubai airshow , where the 777x is expected to \\
    &  dominate with more than 100 orders . \\ 
    \emph{AT:} & now appeals airbus before the dubai airshow, where the 777x with over 100 orders is expected to make \\
    &  the race, directly to the public . \\ 
    \emph{NAT:} & now appeals airbus before the dubai airshow , , where the 777x over over 100 orders is was expected to make make , \\
    &  the race . \\ 
    \emph{LAVA NAT: } & now , airbus is appealing directly to public ahead of the dubai airshow , in which the 777x is expected to \\
    &  dominate with more 100 orders . \\

    \bottomrule\hline
    
    \end{tabular}
    
    \caption{Translating cases on the IWSLT2014 De$\rightarrow$En testset. \emph{AT} represents the Transformer model and \emph{NAT} denotes the model architecture proposed by \citet{jiatao2018nat}.
    \emph{LAVA NAT} denotes the proposed model. Vanilla NAT models suffer from decoding independence problem, which can be solved by the two-way look ahead mechanism. }
    \label{translation-examples}
    \vskip -0.15in
    \end{table*}

\end{document}